\documentclass[letterpaper, 10 pt, conference]{ieeeconf}  

\IEEEoverridecommandlockouts                              

\usepackage{amsmath} 
\usepackage{amssymb}  
\usepackage{booktabs}
\usepackage{graphicx}
\usepackage{caption}
\usepackage{hyperref}
\usepackage{xcolor} 
\usepackage{multirow}
\usepackage{makecell}
\usepackage{float}
\usepackage{subcaption}

\title{\LARGE \bf
VolumeDP: Modeling Volumetric Representation for Manipulation Policy Learning 
}

\author{%
    Tianxing Zhou*$^{1}$,
    Feiyang Xue*$^{1}$,
    Zhangchen Ye*$^{1}$,
    Tianyuan Yuan$^{1,2}$,
    Hang Zhao$^{1,2}$,
    Tao Jiang$^{2}$
    \thanks{$^{1}$IIIS, Tsinghua University, $^{2}$Galaxea AI}
    \thanks{* Equal Contribution.}
    \thanks{Corresponding at: \texttt{ztx23@mails.tsinghua.edu.cn}}
}

\begin{document}

\maketitle
\thispagestyle{empty}
\pagestyle{empty}

\begin{abstract}
Imitation learning is a prominent paradigm for robotic manipulation. However, existing visual imitation methods map 2D image observations directly to 3D action outputs, imposing a 2D–3D mismatch that hinders spatial reasoning and degrades robustness. We present VolumeDP, a policy architecture that restores spatial alignment by explicitly reasoning in 3D. VolumeDP first lifts image features into a Volumetric Representation via cross-attention. It then selects task-relevant voxels with a learnable module and converts them into a compact set of spatial tokens, markedly reducing computation while preserving action-critical geometry. Finally, a multi-token decoder conditions on the entire token set to predict actions, thereby avoiding lossy aggregation that collapses multiple spatial tokens into a single descriptor. 
VolumeDP achieves a state-of-the-art average success rate of $88.8\%$ on the LIBERO simulation benchmark, outperforming the strongest baseline by a substantial $14.8\%$ improvement. It also delivers large performance gains over prior methods on the ManiSkill and LIBERO-Plus benchmarks.
Real-world experiments further demonstrate higher success rates and robust generalization to novel spatial layouts, camera viewpoints, and environment backgrounds. Code and videos are available on the project page: \href{https://yzc0731.github.io/VolumeDP/}{\texttt{yzc0731.github.io/VolumeDP}}.

\end{abstract}

\section{Introduction}
Teaching robots to autonomously accomplish manipulation remains a central research problem. Imitation learning (IL)\cite{pmlr-v164-florence22aImplicitBehavioralCloning} offers a sample-efficient avenue for acquiring diverse motor skills, such as grasping, relocation, and tool use, directly from human demonstrations. Among IL methods, diffusion-based policies have recently shown strong robustness and expressivity. For instance, Diffusion Policy (DP) \cite{DiffusionPolicy} casts action generation as a conditional denoising process, enabling stable learning of high-dimensional, multi-modal action distributions.

Despite these advancements, many policies still rely on 2D visual representations, which struggle with the inherent spatial reasoning required for 3D action outputs. This reliance creates a fundamental 2D-3D misalignment. While some recent works alleviate this misalignment by conditioning policies on explicit 3D inputs (e.g., point clouds) \cite{Ze2024DP3}\cite{gervet2023act3d}\cite{christen2023learning}, this approach necessitates additional, often expensive, depth sensors. In many practical settings, however, RGB cameras are preferable due to their low cost and ability to provide high-resolution appearance and texture cues crucial for precise manipulation.

To overcome the limitations of 2D representations without relying on depth sensors, we propose a novel RGB-only imitation pipeline, called VolumeDP, that integrates a volumetric representation. It consists of three modules. First, a Volumetric Representation module utilizes Volume-Image Cross-Attention to lift the encoded image into volumetric features, capturing critical spatial information. Next, a Spatial Token Generation module learns to distill volumetric features into a set of spatial tokens, enhancing task-relevant features while improving computational efficiency. Finally, conditioned on these spatial tokens, a transformer-based Multi-Token Decoder produces accurate and coherent action sequences.

\begin{figure}[tb]
    \centering
    \includegraphics[width=\linewidth]{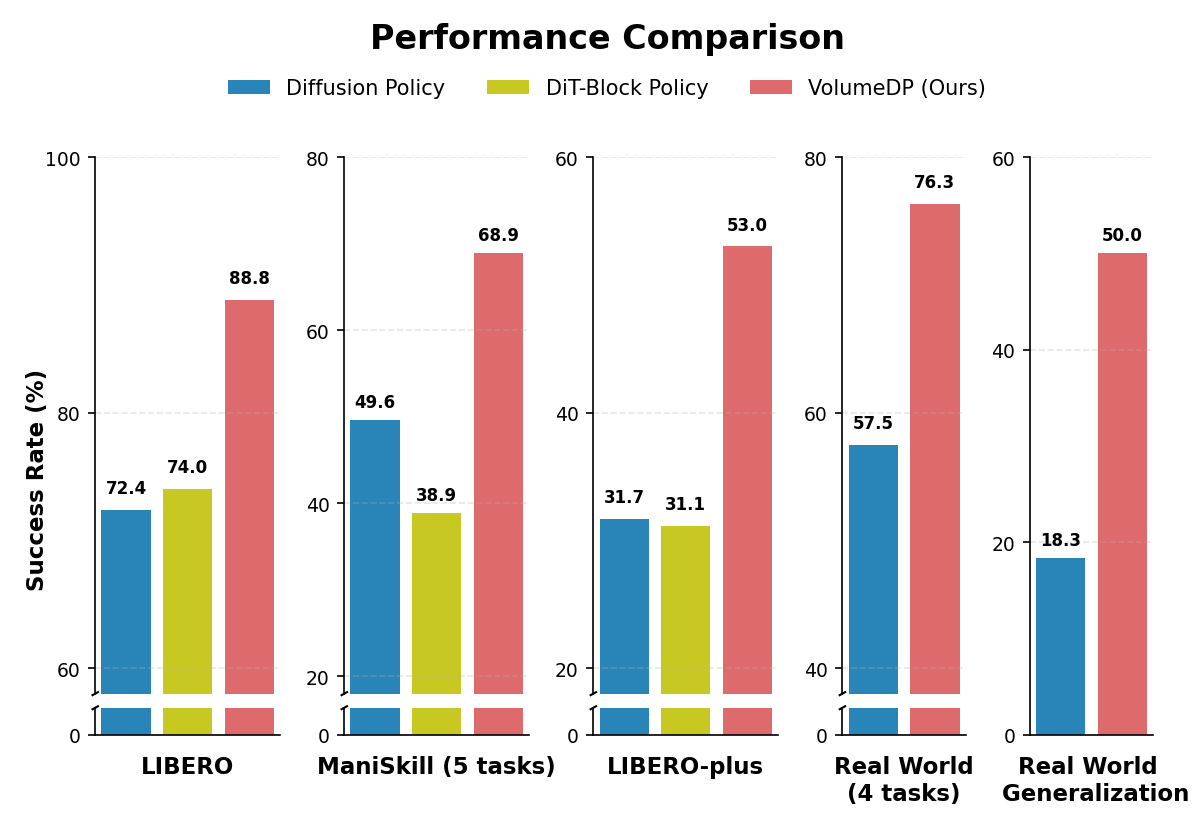}
    \caption{\textbf{VolumeDP} consistently outperforms Diffusion Policy~\cite{DiffusionPolicy} and DiT-Block Policy~\cite{DiTBlockPolicy} by significant margins across \textbf{LIBERO}, \textbf{ManiSkill}, \textbf{LIBERO-plus} simulation benchmarks and \textbf{real-world} manipulation tasks.}
    \label{fig:teaser}
    \vspace{-6mm}
\end{figure}

We demonstrate that our approach substantially outperforms existing baselines across multiple simulation benchmarks and real-world settings, achieving the best performance among the compared methods. Specifically, VolumeDP achieves an $88.8\%$ average success rate on LIBERO~\cite{liu2023libero}, improving upon Diffusion Policy by $16.4\%$ and DiT-Block Policy by $14.8\%$. On ManiSkill and LIBERO-Plus, our method surpasses the strongest baseline on each benchmark by $19.3\%$ and $21.3\%$, respectively.
Real-world experiments further validate these gains, showing higher success rates and robust generalization to novel spatial layouts, camera viewpoints, and background appearances.

Our contributions are summarized as follows:
\begin{itemize}
    \item We introduce \emph{VolumeDP}, an RGB-only imitation policy that resolves the 2D–3D misalignment by constructing a volumetric representation prior to action decoding, thereby enhancing spatial reasoning for complex manipulation.
    \item We design a new architecture: (1) a \emph{Volumetric Representation} built via Volume–Image Cross-Attention; (2) a \emph{Spatial Token Generation} module that distills volumetric features into a compact set of task-relevant spatial tokens while preserving essential features; and (3) a  \emph{Multi-Token Decoder} that conditions on the entire token set to predict accurate, coherent action sequences.
    \item We demonstrate consistent improvements over strong baselines on multiple simulation benchmarks and in real-world experiments, 
    with notably better generalization to variations in robot initial states, camera viewpoints, spatial layouts, and background appearances, highlighting the benefits of a spatially aligned volumetric representation.
\end{itemize}
\section{Related Work}

\subsection{Imitation Learning in Manipulation}

Imitation learning (IL) has emerged as a key paradigm for robotic manipulation, evolving from early behavior cloning (BC) to more expressive and robust frameworks \cite{paraschos2018using}. Traditional BC methods~\cite{ross2010efficient}\cite{ross2011reduction} trained policies to directly map observations to actions, but suffered from covariate shift and limited capacity to model diverse behaviors. This motivated subsequent advances in action representation~\cite{zhou2022domain}. These include modeling actions as parametric distributions such as Gaussian mixtures~\cite{WhatmattersInIL}, using categorical or discrete action spaces to capture multi-modality~\cite{shafiullah2022behavior}, and implicitly modeling actions via latent spaces~\cite{wu2020spatialActionMaps}\cite{chen2024diffusion}. Recently, diffusion models~\cite{ho2020denoising}\cite{song2020score}\cite{songdenoisingDDIM} have gained prominence in IL for their capacity to model complex, high-dimensional, and multi-modal action distribution. Diffusion Policy (DP)~\cite{DiffusionPolicy} exemplifies this trend by conditioning the denoising process on perceptual inputs, enabling policies to iteratively refine noise into coherent actions. Inspired by DP, numerous extensions~\cite{10610175CrosswayDiffusion}\cite{octo}\cite{zhao2025aloha} have demonstrated strong performance across diverse manipulation tasks. However, DP’s U-Net decoder processes only single-token perceptual features, limiting its representational capacity. DiT-Block Policy~\cite{DiTBlockPolicy} addresses this by incorporating adaLN-Zero attention into Transformer-based diffusion decoders, achieving superior performance across a broad range of tasks.

\begin{figure*}[htbp]
    \centering
    \vspace{2mm}
    \includegraphics[width=\textwidth]{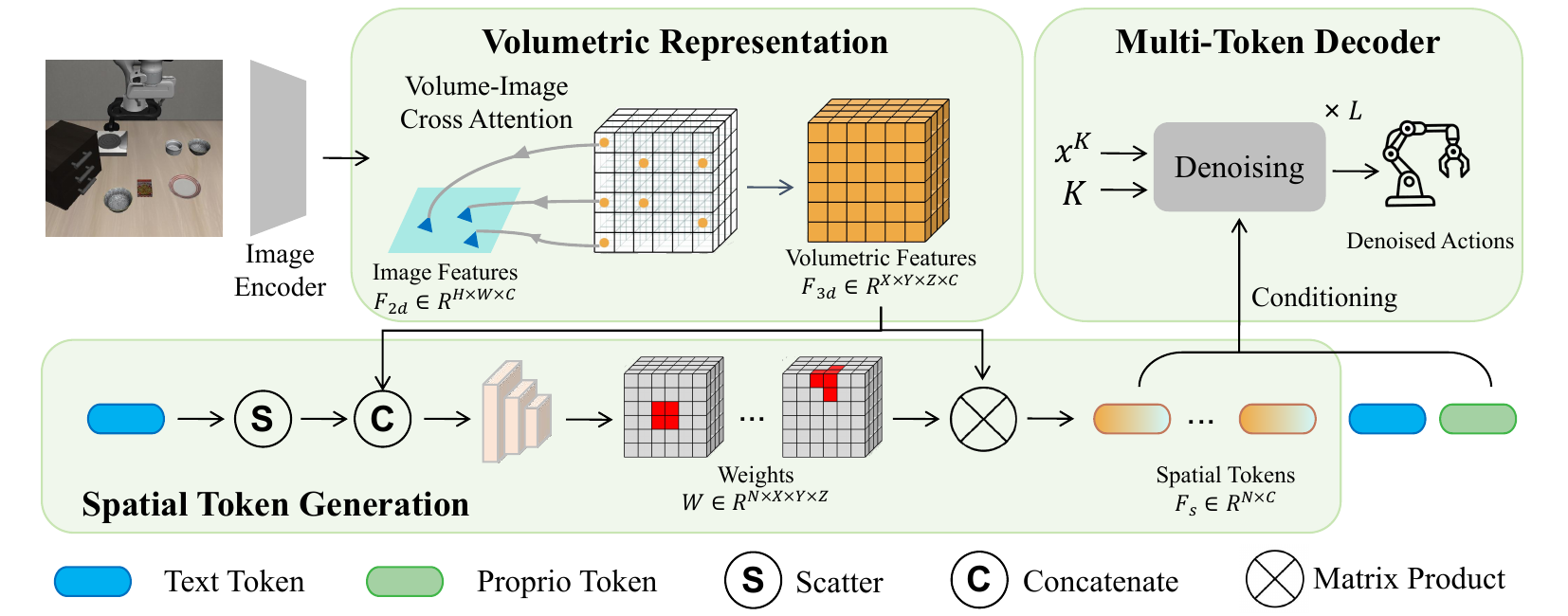}
    \caption{\textbf{Architecture Overview.} Our pipeline consists of three core components: (1) \textbf{Volumetric Representation}: Volume-Image Cross-Attention is applied to construct the Volumetric Representation from image. (2) \textbf{Spatial Token Generation}: Important features are extracted to form spatial tokens. (3) \textbf{Multi-Token Decoder}: The spatial tokens are utilized as conditions for the multi-token denoising decoder.}
    \label{fig:pipeline}
    \vspace{-6mm}
\end{figure*}

\subsection{3D Representation in Manipulation}
Despite advances in diffusion-based manipulation policies, most existing methods rely on 2D representations as conditioning for the denoising decoder. These methods predominantly encode images into 2D feature maps using classical architectures such as ResNet~\cite{DiffusionPolicy} or Vision Transformer~\cite{kim2024openvla}, treating each view independently. This design is inherently limiting, as robots operate in a 3D world where tasks demand strong spatial reasoning. Consequently, 2D representation approaches are prone to overfitting to specific viewpoints and fail to effectively fuse geometric information across views. Recently, 3D Diffusion Policy (DP3)~\cite{Ze2024DP3} has demonstrated that incorporating 3D information like sparse point clouds can significantly enhance model performance, generalization, and data efficiency. Furthermore, other 3D-based policies, including PerAct \cite{pmlr-v205-shridhar23a}, GNFactor \cite{pmlr-v229-ze23a}, and RVT \cite{pmlr-v229-goyal23a}, also underscore the importance of 3D representations in robotic manipulation, enabling advancements in low-dimensional action tasks through improved spatial reasoning.
However, these prior 3D methods typically rely on direct 3D perceptual inputs, such as LiDAR-acquired point clouds or depth image maps, which are expensive to acquire and can be impractical in certain real-world scenarios.

\section{Methodology}
We introduce a novel manipulation pipeline that leverages Volumetric Representation to enhance policy learning. 
As shown in Fig.~\ref{fig:pipeline}, our method comprises three key components. First, our model constructs Volumetric Representation (Sec.~\ref{3Dvolume}) from the image input. Then, the Spatial Token Generation module (Sec.~\ref{tokenselection}) extracts compact information from these volumetric features. Finally, these spatial tokens condition a Multi-Token Decoder (Sec.~\ref{denoise}) for action decoding. 

\subsection{Volumetric Representation}
\label{3Dvolume}
To build the Volumetric Representation, our pipeline encodes input images into image features, defines volumetric space, and then employs Volume-Image Cross-Attention to transform the image features into volumetric representations.

\noindent\textbf{Image Encoding.} Given a sequence of RGB images consisting of the current frame and $T$ past frames,
\(
\mathbf{I} = \{\mathbf{I}_t, \mathbf{I}_{t-1}, \ldots, \mathbf{I}_{t-T}\},
\)
a shared backbone extracts image features \(\mathbf{F}_{2d} \in \mathbb{R}^{H \times W \times C}\) for each image.

\noindent\textbf{Volumetric Space.} We predefine a volumetric space \(\mathbf{Q} \in \mathbb{R}^{X \times Y \times Z}\), corresponding to a cuboid region in the world. 

This volumetric space is parameterized as \( r = [(x_{\text{min}}, y_{\text{min}}, z_{\text{min}}), (x_{\text{max}}, y_{\text{max}}, z_{\text{max}})] \). The bounds \(r\) are task-specific and tunable: they are chosen to span the 3D workspace required by the task, typically covering both the objects of interest and the robot end-effector.
Each voxel in the volume represents a cubic space in the world with a size of $s$ meters. The 3D coordinates $\mathbf{p} = (x, y, z)$ corresponding to a voxel $\mathbf{v} = (i, j, k)$ are obtained as:
\begin{equation}
\label{eq:voxel2location}
    x = i \times s + x_{\text{min}}\quad y = j \times s + y_{\text{min}}\quad z = k \times s + z_{\text{min}}
\end{equation}
Then using the camera extrinsic \(\mathbf{P} \in \mathbb{R}^{3 \times 4}\) and intrinsic \(\mathbf{K} \in \mathbb{R}^{3 \times 3}\), we project these 3D positions $\mathbf{p}$ to image coordinates \((u, v)\), to enable volume-image feature association. 
\begin{equation}
\label{eq:3dto2d}
    \begin{bmatrix}
    u, v, 1
    \end{bmatrix}^\top = \mathbf{K}\mathbf{P} \begin{bmatrix}
    x, y, z, 1
    \end{bmatrix}^\top
\end{equation}

\noindent\textbf{Volume-Image Cross-Attention.} The volumetric features, formulated as \(\mathbf{F}_{3d} \in \mathbb{R}^{X \times Y \times Z \times C}\), are designed to encode both the geometric and semantic information of each voxel.

Direct global attention between all voxels and all image pixels is computationally prohibitive. To address this, we adopt a \emph{deformable attention}~\cite{zhu2021deformabledetr}\cite{BEVFormer}\cite{occ3d}\cite{Cvt-occ} mechanism, which restricts sampling to a small set of learnable offsets around the projected reference point of each voxel.

Specifically, each voxel query \(\mathbf{Q}_v\) is associated with its projected location \(P(v)\) in the image plane via equations \ref{eq:voxel2location} and \ref{eq:3dto2d}. Instead of interpolating a single pixel feature, deformable attention samples a few nearby points with learnable offsets from \(\mathbf{F}_{2d}\) and aggregates them with learned weights. This process is formulated as:
\begin{equation}
\label{eq:crossattn}
    f(\mathbf{Q}_v,\mathbf{F}_{2d})
    = \text{DeformAttn}\!\big(\mathbf{Q}_v,\; P(v),\; \mathbf{F}_{2d}\big)
\end{equation}

Applying this operation across all voxels yields \(\mathbf{F}_{3d}\), a dense, structured representation that supplies fine-grained spatial context for the subsequent modules.

\subsection{Spatial Token Generation}
\label{tokenselection}
In manipulation, most voxels in the volume are empty or irrelevant. The policy should therefore concentrate on the robot end-effector and the manipulated object, which occupy only a small fraction of the scene. Focusing on task-relevant regions also avoids prohibitive computation. To distill salient features while reducing computation cost, we propose the Spatial Token Generation module that compresses the volumetric features into a small set of spatial tokens.

The module takes the volumetric features \(\mathbf{F}_{3d} \in \mathbb{R}^{X \times Y \times Z \times C}\) and a text token \(\mathbf{F}_{\text{text}} \in \mathbb{R}^{C}\), encoded by a text encoder \cite{sanh2019distilbert}.
It first scatters the text token to match the volumetric features' shape and concatenates them to create goal-aware volumetric features \(\mathbf{F}^{g}_{3d} \in \mathbb{R}^{X \times Y \times Z \times (2C)}\). Inspired by TokenLearner~\cite{ryoo2021tokenlearner}, the goal-aware volumetric features are then processed through convolutional layers and normalized using a softmax function to produce a set of weights \(\mathbf{W} \in \mathbb{R}^{N \times X \times Y \times Z}\) within the range \([0,1]\). The weights are used to perform a matrix product with the volumetric features, yielding a set of spatial tokens \(\mathbf{F}_s \in \mathbb{R}^{N \times C}\). Each spatial token represents a weighted sum of the volumetric features according to the weights, effectively distilling the most salient information for the subsequent decoding process. 

To guide the module's focus and accelerate convergence, we introduce auxiliary supervision derived from proprioception, which is readily available in both simulation and real-world settings. For each timestep, we construct a sparse supervision mask around the end-effector's position and around regions where the gripper state changes (open/close), as these typically indicate object interaction. A binary cross-entropy loss encourages high weights in these regions and low weights elsewhere, speeding up learning of the task-relevant areas. 
Notably, both the end-effector pose and gripper state are read directly from the existing proprioceptive signals, requiring no additional annotation.
\begin{table*}[t]
\centering
\vspace{2mm}
\caption{\textbf{LIBERO~\cite{liu2023libero} Results.}We report the success rate for the four task suites in the LIBERO benchmark. * means the baseline is achieved by our implementation. The leading performance is highlighted in \textbf{bold}. 
}
\label{tab:main_results}
\setlength\tabcolsep{8pt}
\fontsize{7}{8}\selectfont
{\renewcommand{\arraystretch}{1.2}
\begin{tabular}[h]{l|cccc|c}
    \toprule
    Method & Libero-Spatial & Libero-Object & Libero-Goal & Libero-Long & Average \\
    \midrule
    BC-Trans~\cite{bai2025rethinking} & $68.0 \pm 1.0$ & $41.8 \pm 1.9$ & $67.8 \pm 10.4$  & $15.8 \pm 2.5$ & 48.4 \\ \hline
    BC-VILT~\cite{bai2025rethinking}& $67.2 \pm 2.3$ & $43.0 \pm 3.9$ & $76.2 \pm 3.0$ & $6.5 \pm 0.9$ & $48.2$ \\ \hline
    Diffusion Policy~\cite{DiffusionPolicy} & $78.3\pm1.1$ & $92.5\pm0.7$ & $68.3\pm1.2$ & $50.5\pm1.3$ & $72.4$ \\ \hline
    Dit-Block Policy*~\cite{DiTBlockPolicy} & $81.1 \pm 2.3$ & $89.8 \pm 0.5$ & $67.6 \pm 0.8$ & $57.6 \pm 4.0$ & $74.0$ \\ \hline
    Ours & $\mathbf{90.7}\pm 1.4$ & $\mathbf{95.9}\pm0.7$ & $\mathbf{89.9}\pm1.2$& $\mathbf{78.7}\pm1.9$& $\mathbf{88.8}_{\textcolor{red}{+14.8}}$  \\
    \bottomrule
\end{tabular}
}
\vspace{-3mm}
\end{table*}

\begin{table*}[t]
\centering
\vspace{2mm}
\caption{\textbf{ManiSkill~\cite{taomaniskill3} results.} We report the success rate for the five tasks from ManiSkill. * means the baseline is achieved by our implementation. The leading performance is highlighted in \textbf{bold}.}
\label{tab:maniskill_results}
\setlength\tabcolsep{8pt}
\fontsize{7}{8}\selectfont
  \begin{tabular}[h]{l|ccccc|c}
    \toprule
    Method & Poke Cube & Place Sphere & Hook Cube & Stack Pyramid & Insert Peg & Average \\
    \midrule
    BC & $26.5 \pm 11.3$ & $6.5 \pm 7.9$ & $2.0 \pm 2.0$  & $0.0 \pm 0.0$ & $0.0 \pm 0.0$ & 7.0 \\ \hline
    Diffusion Policy*~\cite{DiffusionPolicy} & $64.7 \pm 3.1$ & $28.7 \pm 6.1$ & $65.3 \pm 3.1$ & $47.3 \pm 4.6$ & $42.0 \pm 7.2$ & $49.6$ \\ \hline
    Dit-Block Policy*~\cite{DiTBlockPolicy} & $66.0 \pm 5.3$ & $36.7 \pm 1.2$ & $31.0 \pm 4.2$ & $24.0 \pm 12.0$ & $36.7 \pm 5.0$ & $38.9$ \\ \hline
    Ours & $\mathbf{78.0} \pm 8.5$ & $\mathbf{54.7} \pm 1.2$ & $\mathbf{79.0} \pm 1.4$& $\mathbf{57.3} \pm 8.3$& $\mathbf{75.3} \pm 7.0$ & $\mathbf{68.9}_{\textcolor{red}{+19.3}}$\\
    \bottomrule
    \end{tabular}
\vspace{-4mm}
\end{table*}

\begin{table*}[t]
\centering
\vspace{2mm}
\caption{\textbf{LIBERO-Plus~\cite{fei25libero-plus} results.} We report the success rate for the different out-of-distribution(OOD) settings in the LIBERO benchmark to showcase the generalization ability. * means the baseline is achieved by our implementation. The leading performance is highlighted in \textbf{bold}.}
\label{tab:liberoplus_results}
\setlength\tabcolsep{8pt}
\fontsize{7}{8}\selectfont
  \begin{tabular}[h]{l|ccccccc|c}
    \toprule
    Method & Camera & Robot & Language & Light & Background & Noise & Layout & Average \\
    \midrule
    Diffusion Policy*~\cite{DiffusionPolicy} & 1.6 & 32.3 & 77.0 & 25.0 & 19.8 & 20.3 & 42.2 & 31.7 \\ \hline
    Dit-Block Policy*~\cite{DiTBlockPolicy} & 1.6 & 18.3 & 58.1 & 47.5 & 29.2 & 26.8 & 41.2 & 31.1 \\ \hline
    Ours & $\mathbf{23.1}$ & $\mathbf{50.1}$ & $\mathbf{80.8}$ & $\mathbf{56.5}$ & $\mathbf{51.1}$ & $\mathbf{44.7}$ & $\mathbf{66.6}$ & $\mathbf{53.0}_{\textcolor{red}{+21.3}}$ \\ %
    \bottomrule
    \end{tabular}
\vspace{-4mm}
\end{table*}

\subsection{Multi-Token Decoder}
\label{denoise}
To fully leverage the distilled information retained in the spatial tokens extracted from the volumetric features, we introduce the Multi-Token Decoder. This action decoder is formulated as a conditional Denoising Diffusion Probabilistic Model~\cite{ho2020denoising}, which generates the final output through an iterative denoising process. The decoder takes as conditioning the spatial tokens \(\mathbf{F}_s \in \mathbb{R}^{N\times C}\), a proprioceptive token \(z\in\mathbb{R}^{C}\) (from an MLP), and a text token \(\mathbf{F}_{\text{text}}\in\mathbb{R}^{C}\) (from a text encoder).
Given an initial Gaussian noise action \(x^{K}\), a diffusion step index \(k\in\{1,\ldots,K\}\), and a noise predictor \(\epsilon_{\theta}(x^{k},k,\mathbf{F}_s,z,\mathbf{F}_{\text{text}})\), the DDPM update is:
\begin{equation}
\label{eq:ddpm}
    x^{k-1}= \alpha\,\big(x^{k} - \gamma\,\epsilon_{\theta}(x^{k},k,\mathbf{F}_s,z,\mathbf{F}_{\text{text}})\big)\;+\;\mathcal{N}(0, \sigma^2 I)
\end{equation}
where \(\alpha,\gamma,\sigma\) denote the noise schedule coefficients. Starting from \(x^{K}\sim\mathcal{N}(\mathbf{0},\mathbf{I})\), the model iteratively produces \(x^{k-1},\ldots,x^{0}\), with \(x^{0}\) taken as the action output. The training objective follows DP~\cite{DiffusionPolicy}, minimizing the expected prediction error \(\|\epsilon - \epsilon_{\theta}(a+\epsilon,k,\mathbf{F}_s,z,\mathbf{F}_{\text{text}})\|_2^2\) under the forward noising process.

A naïve approach would compress all conditions into a single vector, but this discards cross-token interactions and the rich 3D structure encoded in the spatial tokens. Instead, our decoder preserves the full token set and conditions each transformer block using \emph{adaLN-Zero Attention} block as in the DiT-Block Policy~\cite{DiTBlockPolicy}: the diffusion timestep and conditioning tokens modulate the block via learned scale-and-shift parameters, enabling stable and expressive multi-token conditioning. Given the compatibility of our pipeline with other multi-token injection schemes (e.g., Alternating Condition Injection~\cite{liu2025rdt1b}), we adopt the DiT-Block design in our experiments for its stability and efficiency.

\section{SIMULATION EXPERIMENT}
\subsection{Experimental Setup}
\noindent\textbf{Benchmark.} 
LIBERO~\cite{liu2023libero} is a widely used robotic manipulation benchmark for studying imitation learning~\cite{reuss2024multimodalDiffusionTransformer}\cite{wen2023ATManypointtrajectory}\cite{zheng2026translating}. It comprises four task suites—Spatial, Object, Goal, and Long—each containing 10 tasks, with 50 expert demonstrations per task (500 per suite) for training. We remove all ``no-op" actions and discard demonstrations that fail to complete the task.

\begin{figure}[t]
    \centering
    \vspace{2mm}
    \includegraphics[width=\linewidth]{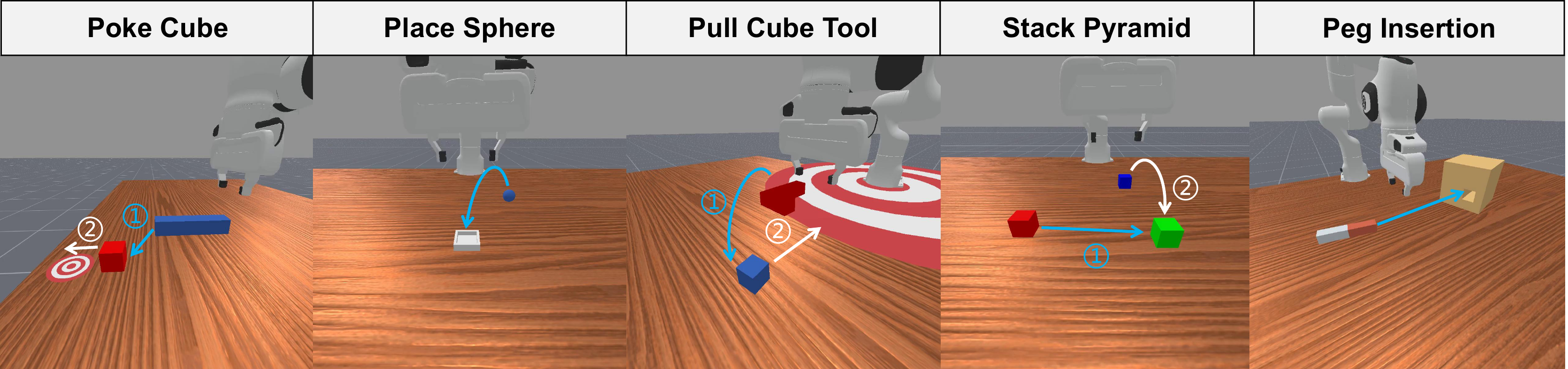}
    \caption{
    \textbf{Visualization of Tasks from ManiSkill.} The arrows indicate the sequential steps required to complete each task. }
    \label{fig:maniskill_tasks}
    \vspace{-3mm}
\end{figure}

In addition to LIBERO, we evaluate our method on five tasks from ManiSkill~\cite{taomaniskill3}: \textit{Poke Cube}, \textit{Place Sphere}, \textit{Hook Cube}, \textit{Stack Pyramid}, and \textit{Insert Peg}. 
This suite presents a diverse set of challenges spanning long-horizon manipulation (e.g., \emph{Stack Pyramid}), precise object operation (e.g., \emph{Insert Peg, Poke Cube}), and complex object interactions (e.g., \emph{Hook Cube}).
Fig.~\ref{fig:maniskill_tasks} illustrates the initial and target state for each task. We collect an average of 150 expert demonstrations from the script policy for each task. 

We further evaluate generalization under distribution shifts using LIBERO-Plus~\cite{fei25libero-plus}, an out-of-distribution (OOD) evaluation protocol built on the LIBERO benchmark. All models are trained on the standard LIBERO dataset and then evaluated under a diverse set of perturbations, including \textit{camera viewpoint}, \textit{initial robot states}, \textit{language instruction}, \textit{lighting}, \textit{background}, \textit{sensor noise}, and \textit{object layout} variations. This benchmark thus provides a rigorous and fair evaluation of model robustness against world-inherent variability.

For all the above benchmarks, we report the success rate (SR) for each experimental task. 
The reported results are averaged over independent runs with three different random seeds.

\noindent\textbf{Implementation Details.} Following DP~\cite{DiffusionPolicy}, we adopt RGB cameras with a fixed resolution of  \(256\times256\) for all experiments.
For LIBERO, only a single third-person RGB camera is used; for ManiSkill, a single third-person RGB camera is used for the \textit{Poke Cube} and \textit{Place Sphere}, and a third-person RGB camera along with a wrist-mounted RGB camera is adopted for the other three tasks. 
At each timestep, the policy receives two consecutive image frames \((t, t\!-\!1)\), the task-specific language instruction, and the current 8D proprioceptive state (including end-effector position, orientation, and gripper status) as input, and outputs a sequence of 7D actions. All methods are uniformly trained and evaluated with 100 denoising steps.

A key parameter in our method is the volumetric space range \(r\). We discretize \(r\) into \(40\times40\times40\) voxels, resulting in a volumetric representation of shape \((H,W,Z,C) = (40,40,40,32)\). This parameter is suite-specific in LIBERO, and task-specific in ManiSkill. Each voxel corresponds to a volume of approximately $1 cm^3$ when mapped to real-world scale. The Spatial Token Generation module produces \(N=200\) spatial tokens, which are projected to 512 dimensions through an MLP before decoding. Consequently, the decoder receives a total of 202 tokens (200 spatial tokens, 1 language token, and 1 proprioceptive token).

\subsection{Main Results}

\noindent\textbf{Baseline.}
We compare our approach against a set of representative baselines to comprehensively evaluate its effectiveness in imitation learning: 
(i) Classical Behavior Cloning (BC); 
(ii) Diffusion Policy (DP)~\cite{DiffusionPolicy}\cite{droid_2024}; and 
(iii) DiT-Block Policy~\cite{DiTBlockPolicy}. 
BC provides a supervised, non-generative reference.  In LIBERO, we adopt results of the widely-used variants BC-Trans and BC-VILT~\cite{liu2023libero}\cite{kim2021vilt} from previous work~\cite{bai2025rethinking}. In ManiSkill, we use the BC implemented by the ManiSkill codebase. 
DP models action sequences via conditional denoising. The DiT-Block Policy employs adaLN-Zero conditioning in the transformer denoising decoder to support multi-token inputs, providing a strong multi-token baseline. Since our decoder is also diffusion-based and multi-token conditioned, these two comparisons reveal the effect of our proposed modules.
Together, these baselines span non-generative and diffusion-based paradigms and enable a fair, direct assessment of VolumeDP.

\noindent\textbf{LIBERO Results.}
Table~\ref{tab:main_results} reports the performance of our method and baselines on the four task suites of the LIBERO benchmark.
Our approach achieves an average success rate of \textbf{88.8\%}, demonstrating clear and consistent gains over existing baselines.

Compared with the Diffusion Policy (DP) baseline, our method yields a remarkable average improvement of \textbf{+16.4\%}, with especially large gains on the challenging LIBERO-Goal ($+21.6\%$) and LIBERO-Long ($+28.2\%$) suites.
These improvements primarily stem from our Volumetric Representation, which encodes structured 3D geometry to enhance spatial perception and reasoning about object layouts.
Additionally, our Multi-Token Decoder facilitates interaction among spatial tokens to exploit rich spatial cues during action prediction.

Furthermore, compared with the competitive DiT-Block Policy, our method achieves a consistent average gain of \textbf{+14.8\%}, including notable improvements on LIBERO-Goal ($+22.3\%$) and LIBERO-Long ($+21.1\%$).
This superiority highlights the effectiveness of combining the Volumetric Representation---which provides fine-grained, spatially-aligned visual features---with the Spatial Token Generation mechanism, which filters out irrelevant information.
Together, they create a highly discriminative and structured representation space for accurate and reliable action generation.

\noindent\textbf{ManiSkill Results.}
Table~\ref{tab:maniskill_results} presents the performance on five selected ManiSkill tasks.
Our method achieves a leading average success rate of \textbf{68.9\%}, outperforming the best baseline by a substantial margin of \textbf{+19.3\%}.
Prominent gains are observed on geometry-sensitive tasks requiring high-precision spatial alignment, such as \textit{Poke Cube} and \textit{Insert Peg}.
This highlights the advantage of explicitly modeling 3D volumetric features for accurate object pose inference and precise action generation. Additionally, our Spatial Token Generation module steers the policy to focus on task-relevant spatial regions, delivering benefits for complex tool-use and object interaction like \textit{Hook Cube}.

\noindent\textbf{LIBERO-Plus Results.} 
Table~\ref{tab:liberoplus_results} also presents the out-of-distribution (OOD) generalization performance on LIBERO-Plus under diverse distribution shifts after standard LIBERO training. 
Our approach achieves a leading average success rate of \textbf{53.0\%}, significantly outperforming Diffusion Policy (31.7\%) and DiT-Block Policy (31.1\%). 
While baselines suffer severe performance degradation or complete collapse under perturbations, our framework maintains robust and reliable performance.

These gains align with VolumeDP's core designs: for \textit{Camera viewpoint}, \textit{Lighting}, and \textit{Sensor noise} perturbations, we improve over Diffusion Policy by +21.5\%, +31.5\%, and +24.4\%, respectively. 
This robustness is attributed to the Volumetric Representation encoding scene geometry in a viewpoint-agnostic 3D space, mitigating viewpoint and illumination sensitivity. 
For \textit{Object layout} (+24.4\%) and \textit{Background} (+31.3\%) variations, the gains stem from the Spatial Token Generation module effectively focusing the policy on manipulable objects and task-critical regions.

\begin{figure}[t]
    \centering
    \includegraphics[width=\linewidth]{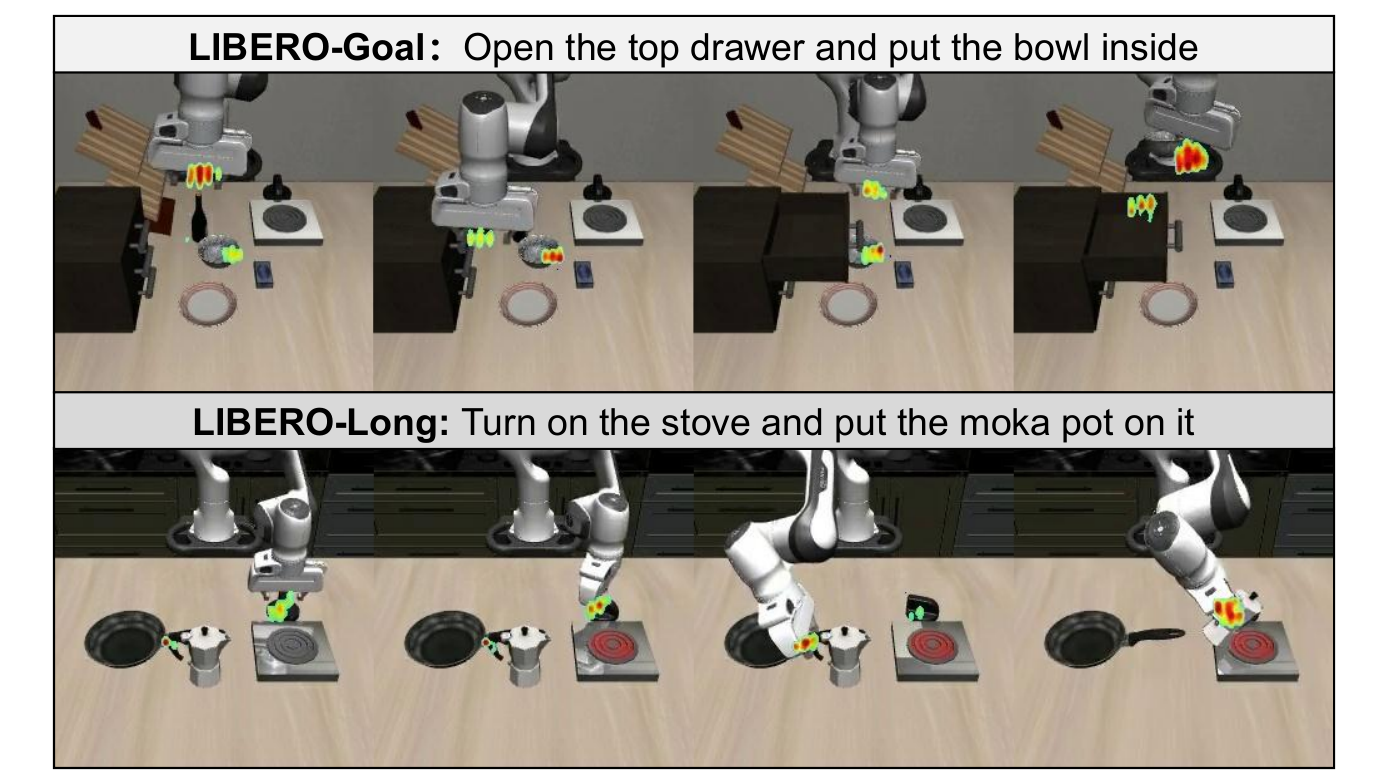}
    \caption{\textbf{Weights Visualization.} Active 3D voxels are projected onto the image plane; brighter red indicates higher activation. The learned weights concentrate on the end effector and the target object, indicating task-relevant spatial awareness.}
    \label{fig:vis}
    \vspace{-8mm}
\end{figure}

\subsection{Ablation Studies}
A series of ablation studies were performed to systematically evaluate the effectiveness of our proposed design. These experiments focus on the three core components: the Volumetric Representation, Spatial Token Generation, and the Multi-Token Decoder. Our findings collectively demonstrate that each component is critical for achieving optimal performance. Specifically, the volumetric representation enhances spatial reasoning, the token generation module efficiently distills information, and the multi-token decoder effectively leverages this data to produce accurate actions.

\noindent\textbf{Ablating Volumetric Representation.} Our 3D volumetric representation is a fundamental element for the model's spatial understanding. To prove its necessity, we replaced it with a standard 2D image feature map \(\mathbf{F}_{2d} \in \mathbb{R}^{X \times Y \times C}\), while keeping the rest of the pipeline unchanged. The output size \(N\) of the Spatial Token Generation module was kept constant for a fair comparison. As shown in Table~\ref{tab:ablation}, the 2D-only model (b) suffered a significant $12.5\%$ drop in success rate on the LIBERO-Spatial benchmark. This result strongly confirms the volumetric representation is indispensable for enabling the model's spatial reasoning capabilities.

\noindent\textbf{Ablating Spatial Token Generation.}
The Spatial Token Generation module is designed to suppress task-irrelevant information and retain action-critical cues in volumetric representation. Qualitatively, Fig.~\ref{fig:vis} shows that the learned weights concentrate on the end effector and the manipulated object, 
and even localize the intended grasp contact region, 
indicating task-relevant spatial focus.

For quantitative analysis, we implemented four variants (c)–(f) in Table~\ref{tab:ablation}, each of which exhibits reduced performance compared to our full model.
(c) ``Random'' randomly samples \(N\) tokens from the volume. 
(d) ``All'' uses all voxels from a coarsened grid of \(10\!\times\!10\!\times\!10\) to keep computation comparable; the performance degradation proves that non-distilled features containing redundant information hinder performance.
(e) ``GT depth'' uses ground-truth depth to unproject pixels in images to the volumetric space and selects \(N\) voxels with the highest point counts. Despite using perfect depth information, it underperforms our learned, goal-aware weights, showing that simple occupancy heuristics are inferior to task-conditioned selection.
(f) ``w/o supervision'' removes the auxiliary supervision on the weights. Although its performance drops, it still outperforms the DP baseline, suggesting that the module can autonomously discover salient regions while the auxiliary signal accelerates and stabilizes convergence.

Taken together, the underperformance of (c)–(f) emphasizes that \emph{learned, goal-aware tokenization} is crucial for aggregating essential information from the volumetric representation.

\noindent\textbf{Ablating Multi-Token Decoder.} After extracting the spatial tokens, effectively utilizing them for action decoding is of great importance. We performed experiments (g) and (h) in Table~\ref{tab:ablation} to validate the necessity of our Multi-Token Decoder.
(g) ``Concat'' concatenates all tokens along the channel dimension and uses a single-token denoising module. Its poor performance proves that compressing multi-token information into a single token leads to significant information loss, hindering performance.
(h) ``Conv'' uses a sequence of convolution layers to process the volumetric features into a single token for the decoder. The performance gap confirms that a simple compressing approach cannot effectively capture and utilize the rich 3D information as our method does.
By outperforming these ablations, our approach confirms that the Multi-Token Decoder is indispensable for fully leveraging the extracted information and achieving strong overall performance.

\begin{table}[t]
\centering
\vspace{2mm}
\caption{\textbf{Ablation results on Libero-Spatial}. Ablating the core components of our pipeline. Replacing each of the parts will cause a performance drop, which demonstrates the efficiency of our model design. }
\label{tab:ablation}
\setlength\tabcolsep{8pt}

\fontsize{7}{8}\selectfont
{\renewcommand{\arraystretch}{1.2}
\begin{tabular}[h]{l|ccc|c}
    \toprule
    Method & \begin{tabular}[c]{@{}c@{}}Volumetric \\ Representation\end{tabular} & \begin{tabular}[c]{@{}c@{}}Spatial Token \\ Generation\end{tabular} & \begin{tabular}[c]{@{}c@{}}Multi-Token \\ Decoder\end{tabular} & \begin{tabular}[c]{@{}c@{}}SR \end{tabular} \\
    \midrule
    (a) & \checkmark  & \checkmark & \checkmark  & 90.7 \\ 
    \midrule
    (b) & \begin{tabular}[c]{@{}c@{}}Image \\ Representation\end{tabular} & \checkmark & \checkmark & 78.2 \\
    \midrule
    (c) & \checkmark & Random & \checkmark  & 62.8 \\
    (d) & \checkmark & All & \checkmark  & 68.4 \\
    (e) & \checkmark & GT Depth & \checkmark  & 75.8 \\
    (f) & \checkmark & w/o supervision& \checkmark & 82.2 \\
    \midrule
    (g) & \checkmark & \checkmark & Concat & 76.4  \\
    (h) & \checkmark & \checkmark & Conv & 80.0 \\
    \bottomrule
\end{tabular}
}
\vspace{-4mm}
\end{table}

\section{REAL WORLD EXPERIMENT}

\begin{figure*}[t]
\centering
\begin{minipage}[t]{0.377\textwidth}
    \vspace{5pt}
    \centering
    \begin{subfigure}[t]{\linewidth}
        \includegraphics[width=\linewidth]{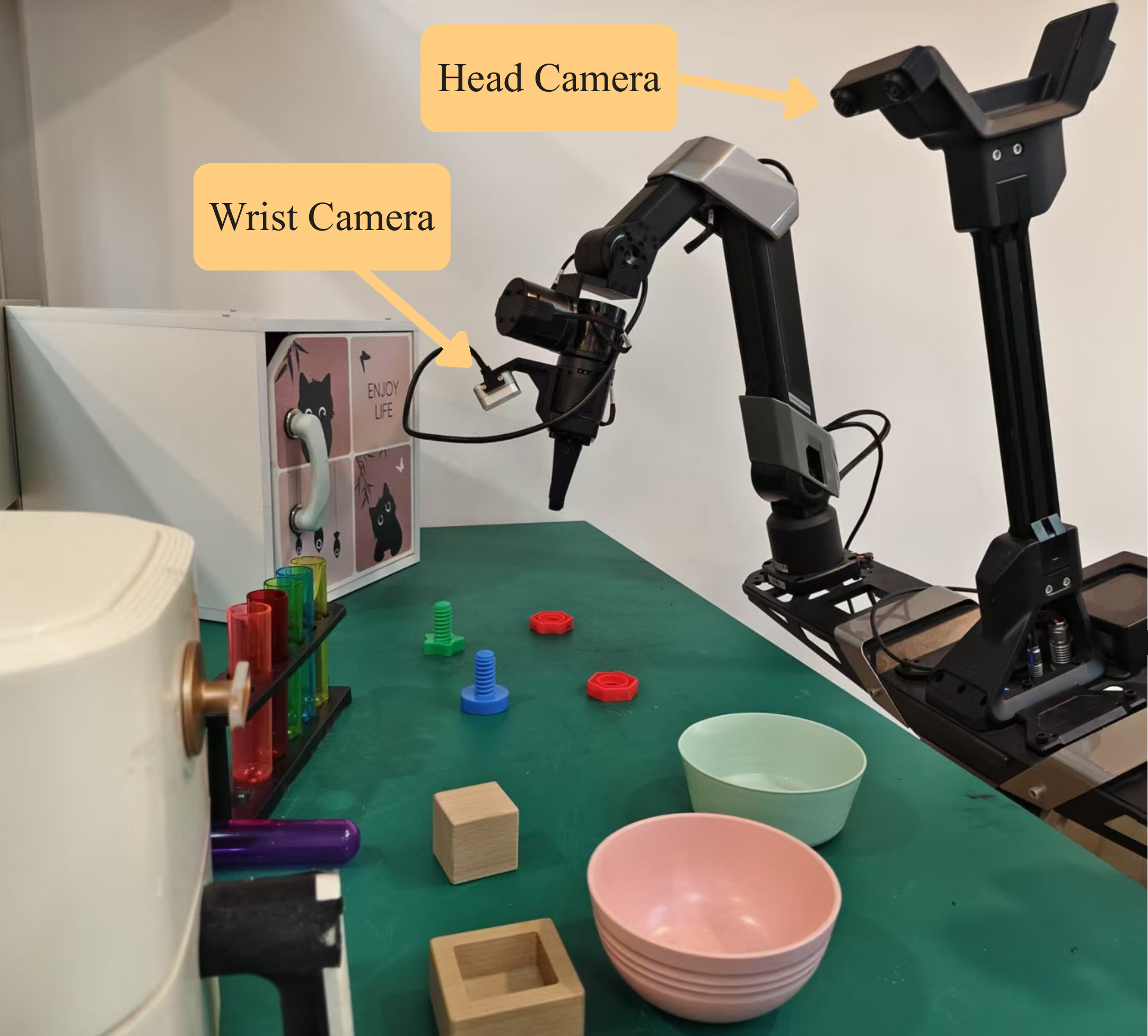}
        \caption{\textbf{Real World Robot.} We highlight the positions of the wrist-mounted and third-person cameras. }
        \label{tab:real_world_picture}
    \end{subfigure}
\end{minipage}
\hfill
\begin{minipage}[t]{0.603\textwidth}
    \vspace{5pt}
    \centering
    \begin{subfigure}[t]{\linewidth}
        \includegraphics[width=0.995\linewidth]{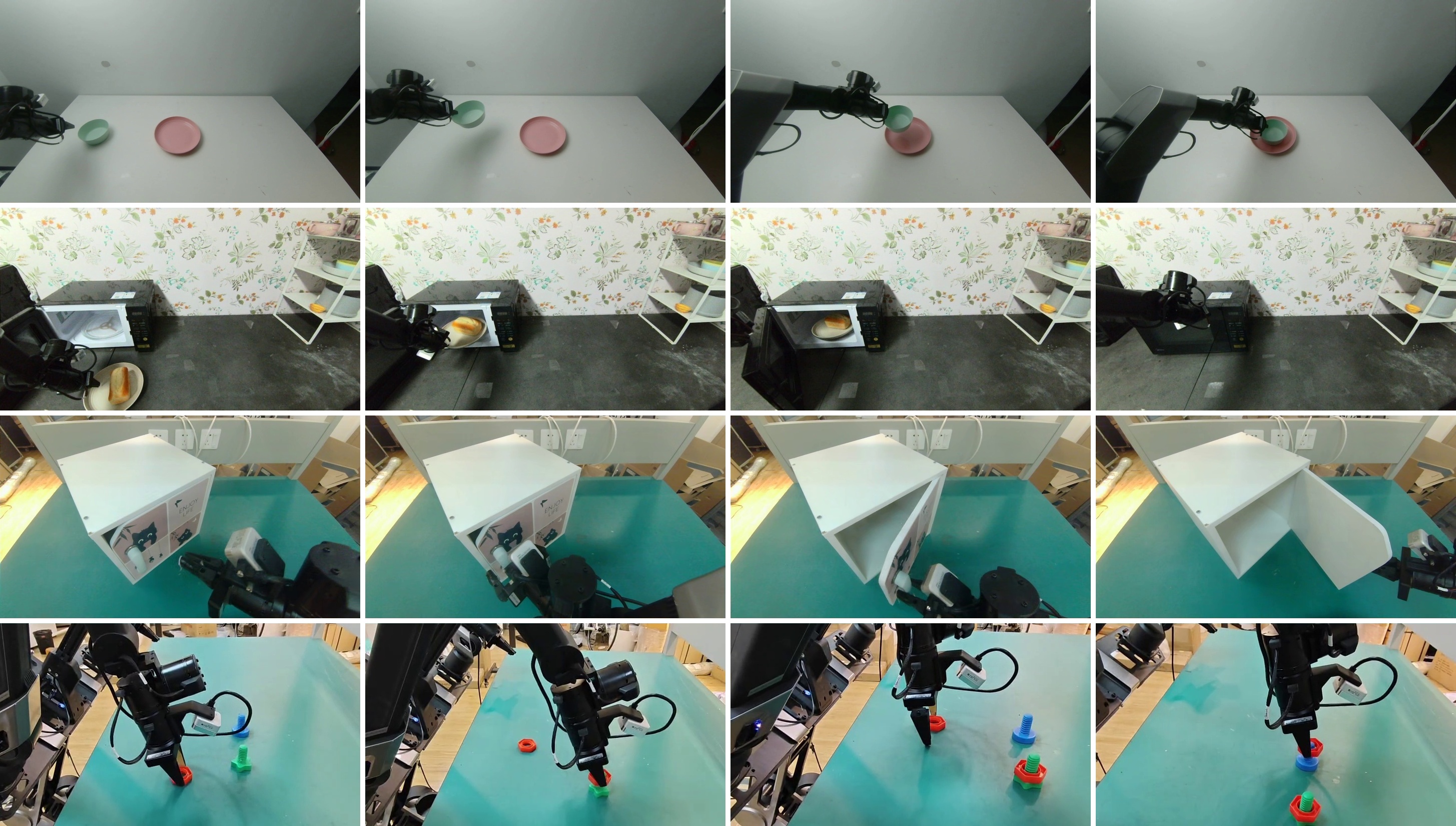}
        \caption{\textbf{Real World Tasks.} To evaluate our model, we run it on four real-world tasks: Placing bowl, Microwave operation, Opening Door and Nut-onto-Screw.}
        \label{fig:pre-train-eval}
    \end{subfigure}
\end{minipage}
\caption{Real World Setup. Left: the real-world robot. Right: the real-world tasks.}
\label{fig:combined_top}
\vspace{-7mm}
\end{figure*}

\subsection{Robot system and Tasks.} 
Our real-world experiments are conducted on Galaxea R1 Lite \cite{Galaxea0}.
The robot features contain two 6-DoF arms and 1-DoF grippers. 
Perception is provided by an RGB head camera and an RGB wrist camera. 
To ensure natural and feasible motions, we adopt an \emph{isomorphic teleoperation} \cite{wu2024gello} scheme that maps human operator movements directly to the robot’s kinematics. For each task, we collect 100 demonstrations by an expert tele-operator. As shown in Fig~\ref{fig:pre-train-eval}, our benchmarks contain the following  four tasks. \textbf{Placing bowl}: The robot is required to pick up a bowl from the table and place it onto a plate. \textbf{Microwave operation}: The robot transfers the plate into the microwave, and then closes the door to initiate heating. \textbf{Door opening}: The robot grasps the handle and pulls the door open. \textbf{Nut onto Screw}: The robot picks up a nut and places it onto a screw. We use only head camera in the first two tasks and use both two cameras in the last two tasks.

\subsection{Analysis}

\noindent\textbf{In-distribution Test.}
We first evaluate the success rate (SR) of our method against the baseline Diffusion Policy (DP) \cite{DiffusionPolicy}. 
Each SR is computed over 20 trials per task.
To evaluate in-distribution performance, the initial object placements during testing are uniformly sampled within the demonstration collection region.

As shown in Table~\ref{tab:real_success_rate}, VolumeDP consistently outperforms DP across all tasks, achieving improvements of +5.0\% SR on \emph{Placing Bowl}, +15.0\% SR on \emph{Microwave Operation}, +20.0\% SR on \emph{Door Opening}, and +35.0\% SR on \emph{Nut onto Screw}, resulting in an average improvement of +18.8\%.

The performance gap becomes more pronounced on tasks such as \emph{Door Opening} and \emph{Nut onto Screw}, which require stronger 3D spatial awareness. 
In particular, the \emph{Nut onto Screw} task demands centimeter-level precise alignment between the nut and the screw, making it highly sensitive to spatial perception errors. 
Similarly, the \emph{Door Opening} task requires accurate estimation of the handle pose, articulation direction, and pulling trajectory.

The \emph{Microwave Operation} task, which involves a longer manipulation sequence, also benefits from improved spatial consistency across multiple interaction stages.

These results suggest that explicit volumetric 3D spatial modeling enables more accurate spatial reasoning and control, which is particularly beneficial for precision-sensitive and contact-rich robotic manipulation tasks.

\begin{table}[t]
\centering
\vspace{1mm}
\caption{\textbf{Real-World Evaluation.} Success rate (SR) over 20 trials on four real-world tasks. Results are reported for both in-distribution (ID) and out-of-distribution (OOD) settings.}
\label{tab:real_success_rate}
\setlength\tabcolsep{2pt}
\fontsize{7.5}{9}\selectfont
{\renewcommand{\arraystretch}{1.2}
\begin{tabular}{l|ccccc|ccc}
\toprule
\multirow{2}{*}{\textbf{Method}}
& \multicolumn{5}{c|}{\textbf{In-distribution}} 
& \multicolumn{3}{c}{\textbf{Out-of-distribution}} \\
\cline{2-9}
& \begin{tabular}[c]{@{}c@{}}Place \\ Bowl \end{tabular} & \begin{tabular}[c]{@{}c@{}} Micro- \\ wave \end{tabular} & \begin{tabular}[c]{@{}c@{}}Open \\ Door \end{tabular} & \begin{tabular}[c]{@{}c@{}}Place \\ Nut \end{tabular} & Avg & Space & View & Env \\
\midrule
Diffusion Policy & 80.0 & 60.0 & 65.0 & 25.0 & 57.5 & 35.0   & 0.0    & 20.0 \\
Ours             & 85.0 & 75.0 & 85.0& 60.0 & 76.3$_{\textcolor{red}{+18.8}}$  & 65.0  & 40.0   & 45.0 \\
\bottomrule
\end{tabular}
}
\vspace{-6mm}
\end{table}

\noindent\textbf{Out-of-distribution Test.}
We further evaluated the generalization ability of the two methods under various distribution shifts, including spatial layout, camera view, and environment background. 

\noindent \textit{Spatial generalization.}
To assess the model's spatial generalization performance, we rearranged the objects to locations different from those seen in the demonstrations. 
 During testing, The target object is placed at out-of-distribution locations (about $10cm$ from the training region).
As shown in Table~\ref{tab:real_success_rate}, DP struggles with spatial displacement, likely due to its reliance on 2D image-space features that are sensitive to object position changes. Consistent with the observations on LIBERO-Plus.

\noindent\textit{View generalization.} To examine the policies' robustness to camera viewpoint changes, we rotated the camera by $\pm5^\circ$ along both the horizontal and vertical directions, resulting in four OOD camera configurations.

Our method demonstrates strong robustness to such minor viewpoint perturbations, whereas the performance of DP drops substantially to 0\% when exposed to unseen viewpoints. This is because 2D feature-based methods are sensitive to changes in the perceived shape and size of objects, and the relative position between objects, which will be highly influenced by the camera view. While our 3D feature representation remains consistent under viewpoint changes due to the viewpoint-invariant nature of the volumetric representation.

\noindent\textit{Environment generalization.}  
We also assessed generalization to environmental variations by modifying background colors, table surfaces, and the appearance of distractor objects during evaluation. Specifically, we created OOD environments by adding different distractor objects, altering wall colors, and changing the table surfaces (e.g., white, black, or blue). As shown in Table~\ref{tab:real_success_rate}, We find that our method is largely insensitive to such environmental changes compared to DP.

Across these shifts, our method consistently outperforms the 2D-based baseline. This improvement arises from our Volumetric Representation and Spatial Token Generation module, which grounds the policy in a 3D space and focuses on task-relevant regions. Consequently, the policy maintains strong robustness to OOD changes.

\section{CONCLUSIONS}
This paper addresses the critical 2D–3D misalignment problem in robotic manipulation by proposing \textit{VolumeDP}, a novel framework that learns spatially grounded policies purely from RGB images. VolumeDP integrates three components: Volumetric Representation, Spatial Token Generation, and Multi-Token Decoder, enabling robust 3D reasoning and spatial understanding. 
With this design, our approach achieves an $88.8\%$ average success rate on LIBERO~\cite{liu2023libero}, improving upon the best baseline method by $14.8\%$, 
while also delivering substantial improvements over existing baselines on the ManiSkill and LIBERO-Plus benchmarks.     
Furthermore, real-world evaluations demonstrate its strong generalization to novel layouts, viewpoints, and environments. VolumeDP provides a practical and scalable paradigm for spatially grounded policy learning, offering promising potential for broader applications requiring advanced spatial perception and reasoning. 

\newpage
\bibliographystyle{IEEEtran}
\bibliography{./IEEEabrv,./IEEEexample}

\begin{thebibliography}{10}
\providecommand{\url}[1]{#1}
\csname url@samestyle\endcsname
\providecommand{\newblock}{\relax}
\providecommand{\bibinfo}[2]{#2}
\providecommand{\BIBentrySTDinterwordspacing}{\spaceskip=0pt\relax}
\providecommand{\BIBentryALTinterwordstretchfactor}{4}
\providecommand{\BIBentryALTinterwordspacing}{\spaceskip=\fontdimen2\font plus
\BIBentryALTinterwordstretchfactor\fontdimen3\font minus \fontdimen4\font\relax}
\providecommand{\BIBforeignlanguage}[2]{{%
\expandafter\ifx\csname l@#1\endcsname\relax
\typeout{** WARNING: IEEEtran.bst: No hyphenation pattern has been}%
\typeout{** loaded for the language `#1'. Using the pattern for}%
\typeout{** the default language instead.}%
\else
\language=\csname l@#1\endcsname
\fi
#2}}
\providecommand{\BIBdecl}{\relax}
\BIBdecl

\bibitem{pmlr-v164-florence22aImplicitBehavioralCloning}
P.~Florence, C.~Lynch, A.~Zeng, O.~A. Ramirez, A.~Wahid, L.~Downs, A.~Wong, J.~Lee, I.~Mordatch, and J.~Tompson, ``Implicit behavioral cloning,'' in \emph{Conference on robot learning}.\hskip 1em plus 0.5em minus 0.4em\relax PMLR, 2022, pp. 158--168.

\bibitem{DiffusionPolicy}
C.~Chi, Z.~Xu, S.~Feng, E.~Cousineau, Y.~Du, B.~Burchfiel, R.~Tedrake, and S.~Song, ``Diffusion policy: Visuomotor policy learning via action diffusion,'' \emph{The International Journal of Robotics Research}, vol.~44, no. 10-11, pp. 1684--1704, 2025.

\bibitem{Ze2024DP3}
Y.~Ze, G.~Zhang, K.~Zhang, C.~Hu, M.~Wang, and H.~Xu, ``3d diffusion policy: Generalizable visuomotor policy learning via simple 3d representations,'' \emph{arXiv preprint arXiv:2403.03954}, 2024.

\bibitem{gervet2023act3d}
T.~Gervet, Z.~Xian, N.~Gkanatsios, and K.~Fragkiadaki, ``Act3d: 3d feature field transformers for multi-task robotic manipulation,'' \emph{arXiv preprint arXiv:2306.17817}, 2023.

\bibitem{christen2023learning}
S.~Christen, W.~Yang, C.~P{\'e}rez-D’Arpino, O.~Hilliges, D.~Fox, and Y.-W. Chao, ``Learning human-to-robot handovers from point clouds,'' in \emph{Proceedings of the IEEE/CVF Conference on Computer Vision and Pattern Recognition}, 2023, pp. 9654--9664.

\bibitem{DiTBlockPolicy}
S.~Dasari, O.~Mees, S.~Zhao, M.~K. Srirama, and S.~Levine, ``The ingredients for robotic diffusion transformers,'' in \emph{2025 IEEE International Conference on Robotics and Automation (ICRA)}.\hskip 1em plus 0.5em minus 0.4em\relax IEEE, 2025, pp. 15\,617--15\,625.

\bibitem{liu2023libero}
B.~Liu, Y.~Zhu, C.~Gao, Y.~Feng, Q.~Liu, Y.~Zhu, and P.~Stone, ``Libero: Benchmarking knowledge transfer for lifelong robot learning,'' \emph{Advances in Neural Information Processing Systems}, vol.~36, pp. 44\,776--44\,791, 2023.

\bibitem{paraschos2018using}
A.~Paraschos, C.~Daniel, J.~Peters, and G.~Neumann, ``Using probabilistic movement primitives in robotics,'' \emph{Autonomous Robots}, vol.~42, no.~3, pp. 529--551, 2018.

\bibitem{ross2010efficient}
S.~Ross and D.~Bagnell, ``Efficient reductions for imitation learning,'' in \emph{Proceedings of the thirteenth international conference on artificial intelligence and statistics}.\hskip 1em plus 0.5em minus 0.4em\relax JMLR Workshop and Conference Proceedings, 2010, pp. 661--668.

\bibitem{ross2011reduction}
S.~Ross, G.~Gordon, and D.~Bagnell, ``A reduction of imitation learning and structured prediction to no-regret online learning,'' in \emph{Proceedings of the fourteenth international conference on artificial intelligence and statistics}.\hskip 1em plus 0.5em minus 0.4em\relax JMLR Workshop and Conference Proceedings, 2011, pp. 627--635.

\bibitem{zhou2022domain}
K.~Zhou, Z.~Liu, Y.~Qiao, T.~Xiang, and C.~C. Loy, ``Domain generalization: A survey,'' \emph{IEEE transactions on pattern analysis and machine intelligence}, vol.~45, no.~4, pp. 4396--4415, 2022.

\bibitem{WhatmattersInIL}
A.~Mandlekar, D.~Xu, J.~Wong, S.~Nasiriany, C.~Wang, R.~Kulkarni, L.~Fei-Fei, S.~Savarese, Y.~Zhu, and R.~Mart{\'\i}n-Mart{\'\i}n, ``What matters in learning from offline human demonstrations for robot manipulation,'' \emph{arXiv preprint arXiv:2108.03298}, 2021.

\bibitem{shafiullah2022behavior}
N.~M. Shafiullah, Z.~Cui, A.~A. Altanzaya, and L.~Pinto, ``Behavior transformers: Cloning $ k $ modes with one stone,'' \emph{Advances in neural information processing systems}, vol.~35, pp. 22\,955--22\,968, 2022.

\bibitem{wu2020spatialActionMaps}
J.~Wu, X.~Sun, A.~Zeng, S.~Song, J.~Lee, S.~Rusinkiewicz, and T.~Funkhouser, ``Spatial action maps for mobile manipulation,'' in \emph{16th Robotics: Science and Systems, RSS 2020}.\hskip 1em plus 0.5em minus 0.4em\relax MIT Press Journals, 2020.

\bibitem{chen2024diffusion}
S.-F. Chen, H.-C. Wang, M.-H. Hsu, C.-M. Lai, and S.-H. Sun, ``Diffusion model-augmented behavioral cloning,'' in \emph{International Conference on Machine Learning}.\hskip 1em plus 0.5em minus 0.4em\relax PMLR, 2024, pp. 7486--7510.

\bibitem{ho2020denoising}
J.~Ho, A.~Jain, and P.~Abbeel, ``Denoising diffusion probabilistic models,'' \emph{Advances in neural information processing systems}, vol.~33, pp. 6840--6851, 2020.

\bibitem{song2020score}
Y.~Song, J.~Sohl-Dickstein, D.~P. Kingma, A.~Kumar, S.~Ermon, and B.~Poole, ``Score-based generative modeling through stochastic differential equations,'' \emph{arXiv preprint arXiv:2011.13456}, 2020.

\bibitem{songdenoisingDDIM}
J.~Song, C.~Meng, and S.~Ermon, ``Denoising diffusion implicit models,'' \emph{arXiv preprint arXiv:2010.02502}, 2020.

\bibitem{10610175CrosswayDiffusion}
X.~Li, V.~Belagali, J.~Shang, and M.~S. Ryoo, ``Crossway diffusion: Improving diffusion-based visuomotor policy via self-supervised learning,'' in \emph{2024 IEEE International Conference on Robotics and Automation (ICRA)}.\hskip 1em plus 0.5em minus 0.4em\relax IEEE, 2024, pp. 16\,841--16\,849.

\bibitem{octo}
O.~Mees, D.~Ghosh, K.~Pertsch, K.~Black, H.~R. Walke, S.~Dasari, J.~Hejna, T.~Kreiman, C.~Xu, J.~Luo \emph{et~al.}, ``Octo: An open-source generalist robot policy,'' in \emph{First Workshop on Vision-Language Models for Navigation and Manipulation at ICRA 2024}, 2024.

\bibitem{zhao2025aloha}
T.~Z. Zhao, J.~Tompson, D.~Driess, P.~Florence, S.~K.~S. Ghasemipour, C.~Finn, and A.~Wahid, ``Aloha unleashed: A simple recipe for robot dexterity,'' in \emph{Conference on Robot Learning}.\hskip 1em plus 0.5em minus 0.4em\relax PMLR, 2025, pp. 1910--1924.

\bibitem{kim2024openvla}
M.~J. Kim, K.~Pertsch, S.~Karamcheti, T.~Xiao, A.~Balakrishna, S.~Nair, R.~Rafailov, E.~Foster, G.~Lam, P.~Sanketi \emph{et~al.}, ``Openvla: An open-source vision-language-action model,'' \emph{arXiv preprint arXiv:2406.09246}, 2024.

\bibitem{pmlr-v205-shridhar23a}
M.~Shridhar, L.~Manuelli, and D.~Fox, ``Perceiver-actor: A multi-task transformer for robotic manipulation,'' in \emph{Conference on Robot Learning}.\hskip 1em plus 0.5em minus 0.4em\relax PMLR, 2023, pp. 785--799.

\bibitem{pmlr-v229-ze23a}
Y.~Ze, G.~Yan, Y.-H. Wu, A.~Macaluso, Y.~Ge, J.~Ye, N.~Hansen, L.~E. Li, and X.~Wang, ``Gnfactor: Multi-task real robot learning with generalizable neural feature fields,'' in \emph{Conference on robot learning}.\hskip 1em plus 0.5em minus 0.4em\relax PMLR, 2023, pp. 284--301.

\bibitem{pmlr-v229-goyal23a}
A.~Goyal, J.~Xu, Y.~Guo, V.~Blukis, Y.-W. Chao, and D.~Fox, ``Rvt: Robotic view transformer for 3d object manipulation,'' in \emph{Conference on Robot Learning}.\hskip 1em plus 0.5em minus 0.4em\relax PMLR, 2023, pp. 694--710.

\bibitem{zhu2021deformabledetr}
X.~Zhu, W.~Su, L.~Lu, B.~Li, X.~Wang, and J.~Dai, ``Deformable detr: Deformable transformers for end-to-end object detection,'' \emph{arXiv preprint arXiv:2010.04159}, 2020.

\bibitem{BEVFormer}
Z.~Li, W.~Wang, H.~Li, E.~Xie, C.~Sima, T.~Lu, Q.~Yu, and J.~Dai, ``Bevformer: learning bird’s-eye-view representation from lidar-camera via spatiotemporal transformers,'' \emph{IEEE Transactions on Pattern Analysis and Machine Intelligence}, vol.~47, no.~3, pp. 2020--2036, 2024.

\bibitem{occ3d}
X.~Tian, T.~Jiang, L.~Yun, Y.~Mao, H.~Yang, Y.~Wang, Y.~Wang, and H.~Zhao, ``Occ3d: A large-scale 3d occupancy prediction benchmark for autonomous driving,'' \emph{Advances in Neural Information Processing Systems}, vol.~36, pp. 64\,318--64\,330, 2023.

\bibitem{Cvt-occ}
Z.~Ye, T.~Jiang, C.~Xu, Y.~Li, and H.~Zhao, ``Cvt-occ: Cost volume temporal fusion for 3d occupancy prediction,'' in \emph{European Conference on Computer Vision}.\hskip 1em plus 0.5em minus 0.4em\relax Springer, 2024, pp. 381--397.

\bibitem{sanh2019distilbert}
V.~Sanh, L.~Debut, J.~Chaumond, and T.~Wolf, ``Distilbert, a distilled version of bert: smaller, faster, cheaper and lighter,'' \emph{arXiv preprint arXiv:1910.01108}, 2019.

\bibitem{ryoo2021tokenlearner}
M.~S. Ryoo, A.~Piergiovanni, A.~Arnab, M.~Dehghani, and A.~Angelova, ``Tokenlearner: What can 8 learned tokens do for images and videos?'' \emph{arXiv preprint arXiv:2106.11297}, 2021.

\bibitem{bai2025rethinking}
S.~Bai, W.~Zhou, P.~Ding, W.~Zhao, D.~Wang, and B.~Chen, ``Rethinking latent redundancy in behavior cloning: An information bottleneck approach for robot manipulation,'' \emph{arXiv preprint arXiv:2502.02853}, 2025.

\bibitem{taomaniskill3}
S.~Tao, F.~Xiang, A.~Shukla, Y.~Qin, X.~Hinrichsen, X.~Yuan, C.~Bao, X.~Lin, Y.~Liu, T.-K. Chan \emph{et~al.}, ``Maniskill3: Gpu parallelized robot simulation and rendering for generalizable embodied ai,'' in \emph{7th Robot Learning Workshop: Towards Robots with Human-Level Abilities}, 2025.

\bibitem{fei25libero-plus}
S.~Fei, S.~Wang, J.~Shi, Z.~Dai, J.~Cai, P.~Qian, L.~Ji, X.~He, S.~Zhang, Z.~Fei, J.~Fu, J.~Gong, and X.~Qiu, ``Libero-plus: In-depth robustness analysis of vision-language-action models,'' \emph{arXiv preprint arXiv:2510.13626}, 2025.

\bibitem{liu2025rdt1b}
S.~Liu, L.~Wu, B.~Li, H.~Tan, H.~Chen, Z.~Wang, K.~Xu, H.~Su, and J.~Zhu, ``Rdt-1b: a diffusion foundation model for bimanual manipulation,'' \emph{arXiv preprint arXiv:2410.07864}, 2024.

\bibitem{reuss2024multimodalDiffusionTransformer}
M.~Reuss, {\"O}.~E. Ya{\u{g}}murlu, F.~Wenzel, and R.~Lioutikov, ``Multimodal diffusion transformer: Learning versatile behavior from multimodal goals,'' \emph{arXiv preprint arXiv:2407.05996}, 2024.

\bibitem{wen2023ATManypointtrajectory}
C.~Wen, X.~Lin, J.~So, K.~Chen, Q.~Dou, Y.~Gao, and P.~Abbeel, ``Any-point trajectory modeling for policy learning,'' \emph{arXiv preprint arXiv:2401.00025}, 2023.

\bibitem{zheng2026translating}
Y.~Zheng, Z.~Ye, W.~Dong, S.~Wang, Y.~Liu, C.~Zhang, C.~Wen, and Y.~Gao, ``Translating flow to policy via hindsight online imitation,'' in \emph{The Fourteenth International Conference on Learning Representations}, 2026.

\bibitem{droid_2024}
A.~Khazatsky, K.~Pertsch, S.~Nair, A.~Balakrishna, S.~Dasari, S.~Karamcheti, S.~Nasiriany, M.~K. Srirama, L.~Y. Chen, K.~Ellis \emph{et~al.}, ``Droid: A large-scale in-the-wild robot manipulation dataset,'' \emph{arXiv preprint arXiv:2403.12945}, 2024.

\bibitem{kim2021vilt}
W.~Kim, B.~Son, and I.~Kim, ``Vilt: Vision-and-language transformer without convolution or region supervision,'' in \emph{International conference on machine learning}.\hskip 1em plus 0.5em minus 0.4em\relax PMLR, 2021, pp. 5583--5594.

\bibitem{Galaxea0}
T.~Jiang, T.~Yuan, Y.~Liu, C.~Lu, J.~Cui, X.~Liu, S.~Cheng, J.~Gao, H.~Xu, and H.~Zhao, ``Galaxea open-world dataset and g0 dual-system vla model,'' \emph{arXiv preprint arXiv:2509.00576}, 2025.

\bibitem{wu2024gello}
P.~Wu, Y.~Shentu, Z.~Yi, X.~Lin, and P.~Abbeel, ``Gello: A general, low-cost, and intuitive teleoperation framework for robot manipulators,'' in \emph{2024 IEEE/RSJ International Conference on Intelligent Robots and Systems (IROS)}.\hskip 1em plus 0.5em minus 0.4em\relax IEEE, 2024, pp. 12\,156--12\,163.

\end{thebibliography}

\end{document}